\documentclass[conference]{IEEEtran}
\IEEEoverridecommandlockouts
\usepackage{cite}
\usepackage{amsmath,amssymb,amsfonts}
\usepackage{algorithmic}
\usepackage{graphicx}
\usepackage{textcomp}
\usepackage{xcolor}
\usepackage{placeins}
\usepackage{float}
\usepackage{stfloats}

\def\BibTeX{{\rm B\kern-.05em{\sc i\kern-.025em b}\kern-.08em
    T\kern-.1667em\lower.7ex\hbox{E}\kern-.125emX}}
\begin{document}

\title{RoPETR: Improving Temporal Camera-Only 3D Detection by Integrating Enhanced Rotary Position Embedding\\
}

\author{\IEEEauthorblockN{Hang Ji}
\IEEEauthorblockA{\textit{Udeer.ai} \\
jihang@udeer.ai}
\and
\IEEEauthorblockN{Tao Ni}
\IEEEauthorblockA{\textit{Udeer.ai} \\
nitao@udeer.ai}
\and
\IEEEauthorblockN{Xufeng Huang}
\IEEEauthorblockA{\textit{Udeer.ai} \\
xufeng@udeer.ai}
\and
\IEEEauthorblockN{Zhan Shi}
\IEEEauthorblockA{\textit{Udeer.ai} \\
shizhan@udeer.ai}
\and
\IEEEauthorblockN{Tao Luo}
\IEEEauthorblockA{\textit{Udeer.ai} \\
luotao@udeer.ai}
\and
\IEEEauthorblockN{Xin Zhan}
\IEEEauthorblockA{\textit{Udeer.ai} \\
zhanxin@udeer.ai}
\and
\IEEEauthorblockN{Junbo Chen *}
\IEEEauthorblockA{\textit{Udeer.ai} \\
junbo@udeer.ai}
}

\maketitle

\begin{abstract}
This technical report introduces a targeted improvement to the StreamPETR\cite{wang2023exploring} framework, specifically aimed at enhancing velocity estimation, a critical factor influencing the overall NuScenes\cite{caesar2020nuscenes} Detection Score (NDS). While StreamPETR exhibits strong 3D bounding box detection performance—as reflected by its high mean Average Precision (mAP)—our analysis identified velocity estimation as a substantial bottleneck when evaluated on the NuScenes dataset. To overcome this limitation, we propose a customized positional embedding strategy tailored to enhance temporal modeling capabilities. Experimental evaluations conducted on the NuScenes test set demonstrate that our improved approach achieves a state-of-the-art NDS of 70.86\% using the ViT-L\cite{wang2023exploring}\cite{dosovitskiy2020image} backbone, setting a new benchmark for camera-only 3D object detection.
\end{abstract}


\section{Introduction}
Camera-only 3D object detection has become increasingly essential for autonomous driving, offering substantial advantages in deployment cost and scalability compared to traditional LiDAR-based methods. By utilizing multi-view camera inputs exclusively, these approaches eliminate dependency on expensive LiDAR sensors while maintaining robust detection performance at longer ranges—a critical requirement for practical driving scenarios. Additionally, vision-based systems inherently align with advanced vision-language models (VLMs), facilitating their integration into next-generation multimodal reasoning and decision-making frameworks.

Recent advancements in Bird's-Eye-View (BEV)\cite{philion2020lift} representation learning, especially those leveraging temporal feature aggregation, have significantly advanced camera-only 3D detection. Techniques like BEVFormer\cite{li2024bevformer} and StreamPETR effectively use temporal fusion to model object movements and scene dynamics, markedly enhancing detection accuracy. Nonetheless, our analysis highlights an underexplored yet critical limitation: current models inadequately balance spatial precision with temporal state estimation, particularly velocity prediction, thereby limiting improvements in the nuScenes Detection Score (NDS). While dense BEV-based detectors achieve high spatial accuracy, their heavy computational burden and reliance on intricate view transformations reduce real-time feasibility. On the other hand, sparse query-based methods suffer from insufficient adaptive aggregation of features across both BEV and image modalities, resulting in notable performance gaps.

Addressing these limitations, position embedding emerges as a crucial component. Although existing methods have attempted to incorporate geometric priors through ray-aware or point-aware spatial embeddings, these approaches remain fragmented. Recognizing this gap, we propose rethinking position embedding as a unified strategy to integrate spatial-temporal reasoning and object-centric feature aggregation. By explicitly modeling geometric correspondences between multi-view inputs and 3D object states, our enhanced positional embedding strategy simultaneously improves temporal coherence and depth-aware representation learning, effectively narrowing the performance gap between mAP and NDS.

In this report, we systematically investigate and optimize positional encoding methods for camera-only 3D detection. Our novel embedding framework unifies object-specific depth priors with temporal geometric constraints, enabling adaptive and efficient feature aggregation across spatial and temporal dimensions. Extensive experiments validate that our approach achieves state-of-the-art results on the nuScenes benchmark, significantly improving both spatial localization and velocity estimation without compromising computational efficiency.

\section{Related Works}
\subsection{Transformer-based 2D Object Detection}
In recent years, object detection has significantly advanced with the introduction of transformers and query-based detection methods\cite{carion2020end}. The transformer architecture, initially designed for modeling long-range dependencies in natural language processing, has been successfully adapted to various computer vision tasks. Its inherent ability to capture global context has enabled the formulation of object detection as a direct set-prediction problem.

DETR\cite{carion2020end} marked a key milestone by introducing learnable queries as representations for objects. In DETR, queries directly interact with image features through a transformer decoder, eliminating traditional steps such as anchor generation and non-maximum suppression (NMS). Although this end-to-end paradigm simplifies detection pipelines, DETR typically suffers from slow convergence due to the complex interplay between queries and global features.

Subsequent works address this limitation by introducing efficient sampling strategies and improved training dynamics. For instance, Deformable DETR\cite{zhu2020deformable} restricts the transformer's attention mechanism to sparse reference points, substantially reducing computational costs and directing attention to more relevant image regions. Similarly, Sparse R-CNN\cite{sun2021sparse} adopts a two-stage framework, initially extracting localized features through ROIAlign\cite{he2017mask} and subsequently refining these features via dynamic convolution, resulting in stabilized and accelerated training convergence.

Further enhancements include adaptive feature mixing and denoising mechanisms. DN-DETR\cite{li2022dn} introduces noisy ground-truth annotations into the decoder, guiding the model to reconstruct accurate bounding boxes, thus improving robustness. Building upon this concept, DINO\cite{zhang2022dino} implements a contrastive denoising strategy and mixed query selection method, significantly boosting convergence speed and overall detection performance.

\subsection{Multi-view 3D Object Detection}
Modern multi-view 3D object detection frameworks typically fall into two paradigms: BEV-based and query-based approaches, distinguished by their strategies for feature aggregation and spatial-temporal modeling.

BEV-based methods rely on the Lift-Splat-Shoot (LSS)\cite{philion2020lift} approach, explicitly constructing dense Bird's-Eye-View representations through depth prediction. BEVDet\cite{huang2021bevdet} first lifts 2D features into 3D voxels using per-pixel depth estimation, facilitating detection directly in BEV space. BEVDepth\cite{li2023bevdepth} further refines geometric accuracy via explicit depth supervision and camera-aware depth estimation, addressing inherent monocular depth biases. Despite their robust spatial localization, these methods are computationally demanding due to dense voxel operations and vulnerability to depth estimation inaccuracies.

Conversely, query-based methods employ learnable queries to perform adaptive feature aggregation. DETR3D\cite{wang2022detr3d} introduces sparse 3D reference points for efficient feature sampling across multiple views. PETR\cite{liu2022petr} innovatively integrates 3D positional embeddings into image features, enabling global interaction between queries and depth-aware features. BEVFormer\cite{li2024bevformer} further enhances this framework by employing grid-shaped BEV queries and deformable attention mechanisms, integrating temporal context via recurrent updates to the BEV states.

\subsection{Temporal Modeling}
Temporal modeling has become critical for enhancing detection robustness and accurate velocity estimation, especially in dynamic driving scenarios. Early approaches primarily targeted short-term feature alignment. BEVDet4D\cite{huang2022bevdet4d}, for example, enhances BEVDet\cite{huang2021bevdet} by compensating for ego-motion and fusing adjacent BEV features, transitioning spatial 3D representations into 4D spatiotemporal contexts. Similarly, PETRv2\cite{liu2023petrv2} temporally aligns 3D positional embeddings across frames, facilitating coherent feature aggregation through transformer attention mechanisms. However, these approaches are limited by their reliance on short temporal windows (typically 2-4 frames), constraining their ability to capture long-term motion dynamics.

To overcome these limitations, recent research explores extended temporal fusion techniques. SOLOFusion\cite{park2022time} proposes an adaptive spatio-temporal trade-off, dynamically balancing feature resolution and historical context duration by maintaining a memory bank of historical features. Although effective, this method suffers from quadratic computational complexity with increased sequence length. StreamPETR\cite{wang2023exploring} addresses these issues by adopting an object-centric temporal propagation strategy, encoding long-term historical information directly into object queries, continuously updating these states online. This framework preserves motion consistency across frames without explicit BEV feature stacking, achieving efficient and robust temporal modeling.

\section{Method}
\subsection{Classical Position Embedding}
Transformers\cite{vaswani2017attention} inherently lack sensitivity to token order and spatial location due to their permutation-invariant nature. To overcome this limitation, positional embeddings are incorporated into token embeddings to explicitly encode positional or spatial information. The two predominant strategies are sinusoidal position embeddings and learnable position embeddings.

Sinusoidal position embeddings, introduced in the original Transformer\cite{vaswani2017attention} model by Vaswani et al., employ fixed sine and cosine functions to generate distinct positional features. Each position's embedding is computed using periodic functions, ensuring uniqueness and enabling generalization beyond the training sequence lengths. A significant advantage of this approach is that it does not introduce additional learnable parameters, keeping the model relatively lightweight.

In contrast, learnable position embeddings involve initializing positional embeddings as parameters optimized during training. This strategy allows embeddings to adapt specifically to dataset characteristics and task requirements. However, the adaptability comes with potential drawbacks, such as limited generalization to positional indices not frequently encountered during training.

\subsection{Position Embedding in 3D Object Detection}
Recent camera-only 3D object detection methods, including PETR\cite{liu2022petr} and 3DPPE\cite{shu20233dppe}, have adapted positional embedding strategies to effectively represent 3D spatial information.

PETR employs a ray-aware encoding method, constructing positional embeddings through a frustum-shaped 3D mesh grid defined by discrete depth candidates. These grid points are projected into LiDAR coordinates using camera intrinsics and extrinsics, followed by encoding via multilayer perceptrons (MLPs).

In contrast, 3DPPE's point-aware encoding explicitly targets depth estimation accuracy. A dedicated DepthNet\cite{cs2018depthnet} predicts pixel-wise depth maps from images, supervised by LiDAR-projected ground-truth depth points. Combining predicted depth with corresponding 2D image coordinates generates precise 3D points, subsequently transformed into LiDAR coordinates for accurate position embedding. Despite improved depth accuracy, pixel-wise depth supervision emphasizes object surface details rather than object-centric locations, creating a mismatch with the query-based detection framework of DETR, which inherently focuses on object centers rather than fragmented surface features.

To mitigate these limitations, we introduce Multimodal Rotary Position Embedding (M-RoPE)\cite{wang2024qwen2}, decomposing rotary embeddings into temporal, height, and width components for unified spatiotemporal encoding. We simplify the 3D embedding by representing the BEV plane as a 2D spatial domain, excluding vertical dimension (Z-axis). Each object's BEV center coordinates $(x,y)$ are normalized within $[0,1]^2$ and multiplied by a logarithmic frequency vector $\omega$ to generate rotation angles $\theta_x$ and $\theta_y$. These angles are applied to channel-pair rotations in query-key ($Q/K$) operations within both self-attention and cross-attention modules, explicitly encoding relative BEV positional offsets.

Further extending the temporal capabilities of M-RoPE, we encode normalized temporal identifiers (frame IDs) $t \in [0,1]$ using a separate frequency vector $\omega_t$, deriving corresponding rotation angles $\theta_t$. Applying these rotational operations effectively captures temporal motion information across frames. StreamPETR's streaming decoder, which naturally propagates object queries through time, benefits significantly from the spatiotemporal rotary embeddings integrated at each timestamp, complementing its inherent motion-aware layer normalization to robustly encode object motion dynamics.

\FloatBarrier
\begin{table*}[!htbp]
\caption{Comparison of other methods on the nuScenes val set.}
\centering
\resizebox{\textwidth}{!}{ 
\begin{tabular}{c|c|c|cc|ccccc}
\hline
{Method} & {Backbone} & {Input Size} & {NDS} & {mAP} & {mATE} & {mASE} & {mAOE} & {mAVE} & {mAAE} \\
\hline

BEVDepth\cite{li2023bevdepth} & ResNet50 & 256 $\times$ 704 & 47.5 & 35.1 & 0.639 & 0.267 & 0.479 & 0.428 & 0.198  \\
SOLOFusion\cite{park2022time} & ResNet50 & 256 $\times$ 704 & 53.4 & 42.7 & 0.567 & 0.274 & 0.511 & 0.252 & 0.181  \\
Sparse4Dv2\cite{lin2023sparse4d} & ResNet50 & 256 $\times$ 704 & 53.8 & 43.9 & 0.598 & 0.270 & 0.475 & 0.282 & 0.179  \\
StreamPETR\cite{wang2023exploring} & ResNet50 & 256 $\times$ 704 & 55.0 & 45.0 & 0.613 & 0.267 & 0.413 & 0.265 & 0.196  \\
\hline
StreamPETR\cite{wang2023exploring} & V2-99 & 320 $\times$ 800 & 57.1 & 48.2 & 0.610 & 0.256 & 0.375 & 0.263 & 0.194  \\
Stream3DPPE\cite{shu20233dppe} & V2-99 & 320 $\times$ 800 & 58.5 & 50.0 & 0.565 & 0.261 & 0.376 & 0.251 & 0.200  \\
RoPETR(Ours) & V2-99 & 320 $\times$ 800 & 61.4 & 52.9 & 0.537 & 0.255 & 0.289 & 0.229 & 0.195  \\
\hline

\end{tabular}
}
\label{tab1}
\end{table*}

\begin{table*}[!htbp]
\caption{Comparison of other methods on the nuScenes test set.}
\centering
\resizebox{\textwidth}{!}{ 
\begin{tabular}{c|c|c|cc|ccccc}
\hline
{Method} & {Backbone} & {Input Size} & {NDS} & {mAP} & {mATE} & {mASE} & {mAOE} & {mAVE} & {mAAE} \\
\hline
DETR3D\cite{wang2022detr3d} & V2-99 & 900 $\times$ 1600 & 47.9 & 41.2 & 0.641 & 0.255 & 0.394 & 0.845 & 0.133  \\
MV2D\cite{wang2023object} & V2-99 & 640 $\times$ 1600 & 51.4 & 46.3 & 0.542 & 0.247 & 0.403 & 0.857 & 0.127  \\
BEVFormer\cite{li2024bevformer} & V2-99 & 900 $\times$ 1600 & 56.9 & 48.1 & 0.582 & 0.256 & 0.375 & 0.378 & 0.126  \\
PETRv2\cite{liu2023petrv2} & V2-99 & 640 $\times$ 1600 & 58.2 & 49.0 & 0.561 & 0.243 & 0.361 & 0.343 & 0.120  \\
BEVDepth\cite{li2023bevdepth} & V2-99 & 640 $\times$ 1600 & 60.0 & 50.3 & 0.445 & 0.245 & 0.378 & 0.320 & 0.126  \\
BEVStereo\cite{li2023bevstereo} & V2-99 & 640 $\times$ 1600 & 61.0 & 52.5 & 0.431 & 0.246 & 0.358 & 0.357 & 0.138  \\
CAPE-T\cite{xiong2023cape} & V2-99 & 640 $\times$ 1600 & 61.0 & 52.5 & 0.503 & 0.242 & 0.361 & 0.306 & 0.114  \\
FB-BEV\cite{li2023fb} & V2-99 & 640 $\times$ 1600 & 62.4 & 53.7 & 0.439 & 0.250 & 0.358 & 0.270 & 0.128  \\
HoP\cite{qiao2022hop} & V2-99 & 640 $\times$ 1600 & 61.2 & 52.8 & 0.491 & 0.242 & 0.332 & 0.343 & 0.109  \\
StreamPETR\cite{wang2023exploring} & V2-99 & 640 $\times$ 1600 & 63.6 & 55.0 & 0.479 & 0.239 & 0.317 & 0.241 & 0.119  \\
\hline

StreamPETR\cite{wang2023exploring} & ViT-L & 640 $\times$ 1600 & 67.6 & 62.0 & 0.470 & 0.241 & 0.258 & 0.236 & 0.134  \\
HoP\cite{qiao2022hop} & ViT-L & 640 $\times$ 1600 & 68.5 & 62.4 & 0.367 & 0.249 & 0.353 & 0.171 & 0.131 \\
RayDN\cite{liu2024ray} & ViT-L & 640 $\times$ 1600 & 68.6 & 63.1 & 0.437 & 0.235 & 0.283 & 0.220 & 0.120 \\
RoPETR(Ours) & ViT-L & 640 $\times$ 1600 & 69.0 & 61.9 & 0.396 & 0.254 & 0.249 & 0.163 & 0.129 \\
RoPETR-e(Ours) & ViT-L & 900 $\times$ 1600 & 70.9 & 64.8 & 0.379 & 0.227 & 0.248 & 0.173 & 0.125 \\
\hline

\end{tabular}
}
\label{tab1}
\end{table*}

\section{Experiments}
\subsection{Dataset and Metrics}
We evaluate our approach on the large-scale nuScenes dataset.
The nuScenes dataset comprises 1000 video sequences, with each sequence lasting approximately 20 seconds. The dataset is split into 700 training videos, 150 validation videos, and 150 testing videos, with annotations provided at a rate of 2 Hz (i.e., one annotation every 0.5 seconds). This dataset features up to 1.4 million annotated 3D bounding boxes covering ten common object classes.

\subsection{Implementation Details}
Our proposed method builds upon the StreamPETR framework, closely following its original configuration. We perform extensive experiments using the nuScenes dataset, adopting the V2-99\cite{lee2020centermask} backbone initialized from the DD3D\cite{park2021pseudo} checkpoint for validation experiments. To further enhance performance on the test set, we scale our method by utilizing the more powerful ViT-L\cite{wang2023exploring}\cite{dosovitskiy2020image} backbone.

During training, we employ the same streaming video training strategy as StreamPETR, training models for 24 epochs using a cosine annealing learning rate schedule. All experiments are conducted without utilizing the Class-Balanced Grouping and Sampling (CBGS)\cite{zhu2019class} strategy, and training is carried out on 8×A100 GPUs.

Additionally, we incorporate the RayDN\cite{liu2024ray} module into our framework to further enhance detection performance.

\subsection{Main Results}
On the nuScenes validation set, the baseline StreamPETR model with a ResNet50\cite{he2016deep} backbone at 256×704 resolution achieves an NDS of 55.0\% and an mAP of 45.0\%. Upgrading the backbone to V2-99\cite{lee2020centermask} and increasing the resolution to 320×800 notably boosts the performance, improving StreamPETR's NDS to 57.1\% and mAP to 48.2\%. By incorporating the point-aware positional embedding from 3DPPE\cite{shu20233dppe}, the enhanced Stream3DPPE model achieves a higher NDS of 58.5\% and an mAP of 50.0\%. Our proposed RoPETR method significantly outperforms these variants, achieving an NDS of 61.4\% and mAP of 52.9\%, clearly demonstrating the efficacy of our rotational positional encoding approach.

Consistent performance improvements are observed on the nuScenes test set. Prior methods such as DETR3D\cite{wang2022detr3d} and MV2D\cite{wang2023object} reported moderate results of 47.9\% and 51.4\% NDS, respectively. StreamPETR\cite{wang2023exploring}, equipped with a V2-99 backbone at a resolution of 640×1600, attains an NDS of 63.6\% and mAP of 55.0\%. Employing the stronger ViT-L\cite{wang2023exploring}\cite{park2022time} backbone, StreamPETR's performance further improves, reaching 67.6\% NDS and 62.0\% mAP. Our refined RoPETR model achieves superior performance, with an NDS of 69.0\% and mAP of 61.9\%. To fully leverage our method’s potential, we introduce RoPETR-e, an enhanced variant utilizing a higher input resolution (900×1600) combined with Test-Time Augmentation (TTA). RoPETR-e sets a new state-of-the-art benchmark on the nuScenes test set, achieving exceptional results of 70.9\% NDS and 64.8\% mAP.

\section{Conclusion}

In this technical report, we presented RoPETR, an advanced position embedding method that significantly improves camera-only 3D object detection, particularly addressing the challenging task of velocity estimation in autonomous driving scenarios. Our approach innovatively combines object-centric positional encoding with a multimodal rotary embedding strategy, effectively capturing spatiotemporal dependencies and motion dynamics.

\bibliography{references}
\bibliographystyle{plain}

\end{document}